\documentclass{article}

\usepackage{PRIMEarxiv}

\usepackage[utf8]{inputenc} 
\usepackage[T1]{fontenc}    
\usepackage{hyperref}       
\usepackage{url}            
\usepackage{booktabs}       
\usepackage{amsfonts}       
\usepackage{nicefrac}       
\usepackage{microtype}      
\usepackage{lipsum}
\usepackage{fancyhdr}       
\usepackage{graphicx}       
\graphicspath{{images/}}    
\usepackage{titlesec}
\usepackage{caption}
\usepackage{array}
\usepackage{booktabs}
\usepackage{longtable}
\usepackage{multirow}
\usepackage{tablefootnote}
\usepackage{float}

\usepackage[labelfont=bf]{caption}
\newcolumntype{C}[1]{>{\centering\arraybackslash}p{#1}}

\titleformat{\paragraph}
{\normalfont\normalsize\bfseries}{\theparagraph}{1em}{}
\titlespacing*{\paragraph}
{0pt}{3.25ex plus 1ex minus .2ex}{1.5ex plus .2ex}

\newenvironment{conditions}
  {\par\vspace{\abovedisplayskip}\noindent\begin{tabular}{>{$}l<{$} @{${}={}$} l}}
  {\end{tabular}\par\vspace{\belowdisplayskip}}

\title{LLM-Assisted Content Analysis: Using Large Language Models to Support Deductive Coding
}

\author{
  Robert Chew, John Bollenbacher, Michael Wenger \\
  Center for Data Science and AI \\
  RTI International \\
  \texttt{\{rchew, jmbollenbacher, mwenger\}@rti.org} \\
   \And
  Jessica Speer, Annice Kim \\
  Center for Communication and Media Impact \\
  RTI International \\
  \texttt{\{jlspeer, akim\}@rti.org} \\
}

\begin{document}
\maketitle

\begin{abstract}
Deductive coding is a widely used qualitative research method for determining the prevalence of themes across documents.  While useful, deductive coding is often burdensome and time consuming since it requires researchers to read, interpret, and reliably categorize a large body of unstructured text documents.  Large language models (LLMs), like ChatGPT, are a class of quickly evolving AI tools that can perform a range of natural language processing and reasoning tasks.   In this study, we explore the use of LLMs to reduce the time it takes for deductive coding while retaining the flexibility of a traditional content analysis.  We outline the proposed approach, called \emph{LLM-assisted content analysis} (LACA), along with an in-depth case study using GPT-3.5 for LACA on a publicly available deductive coding data set. Additionally, we conduct an empirical benchmark using LACA on 4 publicly available data sets to assess the broader question of how well GPT-3.5 performs across a range of deductive coding tasks. Overall, we find that GPT-3.5 can often perform deductive coding at levels of agreement comparable to human coders. Additionally, we demonstrate that LACA can help refine prompts for deductive coding, identify codes for which an LLM is randomly guessing, and help assess when to use LLMs vs. human coders for deductive coding. We conclude with several implications for future practice of deductive coding and related research methods.
\end{abstract}


\section{Introduction}
Content analysis is widely used in qualitative research to analyze and interpret the characteristics of text, or other forms of communication, due to its systematic and unobtrusive nature \cite{united_states_government_accountability_office_program_evaluation_and_methodology_division_content_1996}.  Content analysis typically involves selecting a sample of text data, defining categories to classify the content, and then coding the content according to the  categories with definitions. This is typically referred to as deductive coding in which researchers develop a coding scheme based on existing theories and research prior to the coding process.  This is in contrast with inductive coding which involves not defining categories \emph{a priori}, but rather identifying and naming  categories that emerge from the text during the coding process. While more rigid, deductive coding is  more well suited for generalizing results across studies \cite{linneberg2019coding}.

Despite its strengths, deductive coding is a time-consuming process, particularly when coding substantial amounts of data \cite{strauss_qualitative_1987} and for topics that may be nuanced or infrequently mentioned. Coding requires researchers to carefully read and code each piece of content, possibly multiple times, to ensure that they are accurately capturing all relevant information, properly interpreting the text, and applying the category definitions faithfully.  This burden becomes magnified when developing and refining the codebook, training coders, and measuring inter-rater reliability to ensure code definitions are well-defined and can be coded consistently \cite{neuendorf_content_2017}. 

Recently, generative large language models (LLMs) \cite{wei2021finetuned, brown_language_2020} have demonstrated remarkable progress toward achieving human-level performance on a number of natural language understanding and reasoning tasks \cite{liang2022holistic}.  For example, the developers of GPT-4 claim that it performs in the 90th percentile on the Uniform Bar Exam and 93rd and 89th percentiles for the SAT Reading and SAT Math exams respectively, despite not being trained to solve them specifically \cite{openai_gpt-4_2023}. 

In this study, we explore the potential of LLMs for conducting qualitative coding tasks. Specifically, we propose a methodology for incorporating LLMs into deductive coding, called \emph{LLM-assisted content analysis} (LACA), that aligns with traditional content analysis and provides a way to assess when to use LLMs vs. human coders. We demonstrate this methodology with an in-depth case study, as well as across four diverse publicly available data sets, highlighting notable differences in performance across document types, codebooks, and code categories.

\subsection{Background and Related Work}
There is a long history in the qualitative research literature of using computers to help perform content analysis on text \cite{weber_computer-aided_1984, roberts2020text}.  Proposed as early as 1966, computer-assisted content analysis \cite{stone_general_1966} uses pre-defined dictionaries of key terms and phrases related to the phenomena under study and searches for them within a set of documents.  Term frequencies can then be calculated, grouped, and compared to each other or across documents.  Computer-assisted content analysis benefits from being explicit and transparent in coding decisions since codes are assigned based solely on the presence of terms.  However, for dictionary-based methods to work well, they must be constructed carefully to capture the intended semantic meanings of words within the study domain. For example, Loughran and McDonald \cite{loughran_when_2011} used the standard Harvard-IV-4 dictionary to analyze the sentiment of corporate earnings reports. They note that many terms assigned negative connotations in the dictionary (e.g., “crude”, “tax”, “cost”) have positive or neutral connotations when used in the context of the 10-K filings (e.g., use of “crude oil” by oil and gas companies).  Additionally, constructing new custom dictionaries can be burdensome and can require substantial validation to ensure included terms sufficiently measure the underlying construct of interest.

To overcome computer-assisted content analysis’s disadvantages, researchers proposed using supervised machine learning and natural language processing (NLP) to support deductive coding. Unlike a dictionary approach, where researchers define how to code documents by creating term sets for each category of interest, supervised learning methods learn how to code documents from prior examples of documents both containing and not containing the category of interest.  Since these models are trained to distinguish between categories within the corpus under study, they can often result in high quality automated coding with fewer concerns of the semantic drift associated with using an off-the-shelf dictionary designed for a different context.  Researchers have used supervised learning models to support deductive coding under various settings, including thematic analysis of German news articles \cite{scharkow_thematic_2013}, interpretation of U.S Supreme Court amicus briefs \cite{evans2007recounting}, and coding of open-ended survey responses \cite{baumgartner_framework_2021}.  However, supervised machine learning has historically seen limited adoption in content analysis due to there being insufficient labeled data to obtain strong results, especially for rare categories that may be of most interest to social scientists. Additionally, without knowing why model decisions are made, social scientists may be reluctant to trust machine learning predictions \cite{chen2016challenges}.

Recently, generative LLMs such as ChatGPT are being used for tasks similar to deductive coding.  The advantages of these models are that they can often produce compelling results for zero-shot learning (no examples) or few-shot learning (few examples), addressing a limitation of traditional supervised learning.  Gilardi et al. \cite{gilardi_chatgpt_2023} use ChatGPT to perform several annotation tasks (relevance, stance, topics, and frame detection) and compared its performance to crowd-workers recruited from Amazon Mechanical Turk.  They found that ChatGPT’s zero-shot accuracy exceeded crowd-workers for all tasks but topic categorization.  Similarly, Tornberg \cite{tornberg_chatgpt-4_2023} compared the performance of zero-shot GPT-4 annotation to experts and crowd workers to determine whether tweets come from U.S. Democrat or Republican politicians. He found that GPT-4 outperforms crowd-workers and experts in accuracy and inter-rater reliability.  Closest to our work, Xiao et al., \cite{xiao_supporting_2023} show early results that LLMs may hold promise for deductive coding.  They used an early version of GPT-3.5 (text-davinci-002) to perform deductive coding on children’s curiosity-driven questions, comparing two different prompt designs:  one that includes exemplar coding decisions within the prompt and another that modifies an existing codebook into a prompt.  They found that the codebook-based prompts outperformed the example-based prompts, though both underperformed when compared to the expert coders. 

Our contribution extends previous work by proposing a holistic role for LLMs within the larger deductive coding framework.  Particularly, we focus on prior critiques of applying machine learning to qualitative coding that encompass both the burden of creating training data and the need for model reasoning in coding decisions.  Additionally, our work contains a benchmark study like previous works that compares how well a particular LLM performs on a set of deductive coding tasks relative to human coders \cite{gilardi_chatgpt_2023,tornberg_chatgpt-4_2023,xiao_supporting_2023}.  This component serves a dual purpose of demonstrating how LACA would apply in different settings while also contributing to the growing literature on how well current generation of LLMs perform on qualitative coding tasks.

\section{Methods}
\label{sec:methods}

\subsection{LLM-Assisted Content Analysis (LACA)}
\begin{figure}
    \includegraphics[width=\textwidth]{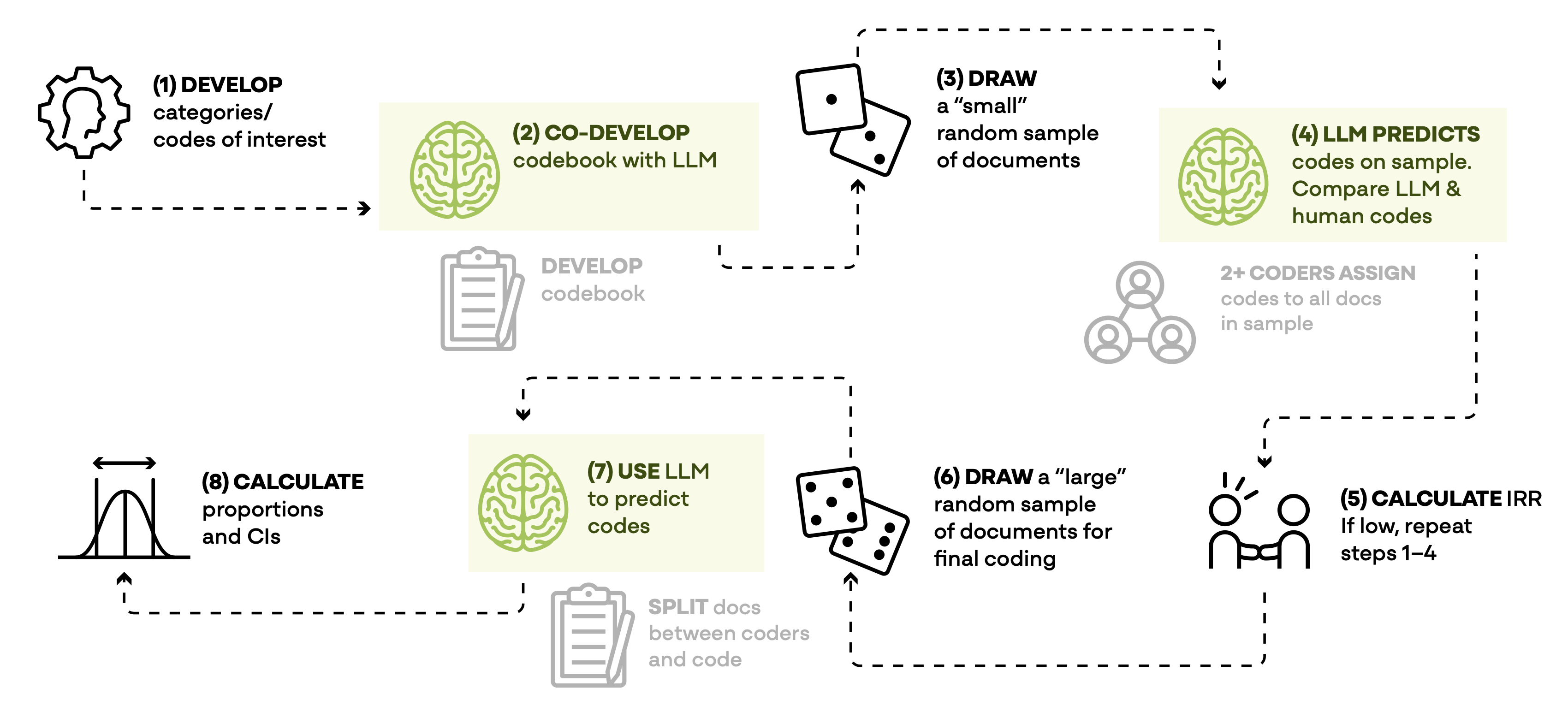}
    \caption{LLM-Assisted Content Analysis (LACA) Process Diagram}
    \label{fig:process_diagram}
\end{figure}

LACA builds off Neuendorf’s approach to content analysis \cite{neuendorf_content_2017} which emphasizes how concepts from survey research, such as measurement, reliability, and population inference, can help qualitative researchers systematically conduct a content analysis.  First, we outline the steps of a deductive content analysis following this approach.  Next, we describe LACA, highlighting components in which it differs from the typical deductive coding workflow (Figure \ref{fig:process_diagram}).

Typically, the first step in a content analysis is to develop measures (categories) that operationalize the concepts under study.  For deductive coding, these measures should be drawn from theory when possible and allow researchers to test hypotheses about the content.  A key component of measure development is creating a coding scheme and accompanying codebook.  A codebook is document that contains guidelines for assigning codes, often containing definitions of categories, instructions for assigning categories, and examples of coded data.  Once the researcher has determined which codes are necessary and developed instructions for how to code them, a sample of documents is drawn to start coding, ideally at random to support downstream statistical inference.  This assumes that coding a census of all applicable documents is normally infeasible.  To ensure that the measures can be reliably understood and consistently applied to documents, two or more coders code all the documents in the sample and inter-rater reliability (IRR) metrics are calculated to quantify the agreement between coders.  If the IRR metrics are sufficiently high, a much larger random sample of documents are selected for final coding.  This larger sample is often necessary if a primary goal of the analysis is to quantify how often codes occur (or co-occur with one another) reliably while considering the impact of sampling variability.  Once the final sample is coded, point estimates such as counts and proportions can be calculated along with their associated confidence intervals.  

LACA differs from this typical workflow in three components: (1) codebook co-development with an LLM and tests of validity; (2) tests of reliability between human coders and the LLM; and (3) replacement of manual coding with LLM coding for the final coded data set.

\subsubsection{Codebook Co-Development and Tests of Validity}
The first step in LACA is to use LLMs when developing the codebook.  Like a human coder, an LLM will need to use the codebook as instructions (i.e., prompt) for how to code documents. 
Codebook co-development with LLMs requires testing the codebook against an LLM on a sample of uncoded documents.  The goal of this assessment is to help determine if the LLM “understands” the coding task and can produce valid output that captures the underlying construct of interest.  In addition to manual reviewing the LLM-assigned coding decisions, we propose two methods to assist in this assessment: (1) conducting hypothesis tests to assess if the LLM generated coding decisions can be distinguished from random noise, and (2) using prompts that not only provide a coding decision, but also a reason for the coding decision. 

\paragraph{Tests of Randomness}
Our theory motivating hypothesis testing is that if the model severely does not understand the concept that is supposed to code for, it will be forced to assign codes effectively at random.  For codes requiring a “Yes” / “No” decision, we would expect a model that is guessing to produce a distribution of coding decisions that’s approximately an even split between “Yes” and “No”.  Hypothesis testing can be used to assess if the frequency of model generated code decisions is consistent with this behavior, helping identify coding tasks early on which the model may not understand and consistently underperforms. Details on tests are in provided in Section \ref{sec:eval} and evidence of their applicability is presented in Sections \ref{sec:tests-random} and \ref{sec:summary}.  However, one drawback of these tests is that they will reject all coding decision processes whose frequency mimics a uniform distribution, regardless if the model is actually randomly guessing.  Because of this, manually reviewing individual observations with the method outlined below is still highly recommended.

\paragraph{Reasons for Coding Decisions}
To help assess face validity of the model generated codes, we use prompts that not only provide a coding decision, but also a reason for the coding decision. Prior work suggests that adding reasons to prompts can help annotators better understand the model’s decisions \cite{elsherief_latent_2021, zhu_can_2023}.  Additionally, similar work that force LLMs to step through their reasoning for complex tasks has been shown to help improve task performance \cite{wei_chain--thought_2023}.  This step in LACA is included to help address the issue of trusting LLMs, since it can, in theory, help assess the quality of coding at the individual observation level.  However, caution should be taken to not over-interpret LLM-generated responses, since it is not well understood to what extent model generated reasons actually reflect how a model makes coding decisions.  We consider the model justifications primarily as an aid for human reviewers, as opposed to a faithful mechanistic account of the model's inner workings.  Instructive examples of model code reasoning are provided in Sections \ref{code-reasons} and \ref{code-reasons-cal}.

\paragraph{Model Assessment and Codebook Revision}
After running hypothesis tests and reviewing the model generated codes and reasons, it may be apparent that some codes are underperforming, either because the model does not understand the coding task or because the task its coding for differs from the concept in mind.  The latter issue may be due to an underspecified codebook and may be addressed by providing more context in the codebook descriptions or providing additional examples \cite{xiao_supporting_2023}.  Ideally, codebooks provide definitions, instructions, and examples that allow a human coder who is new to the task enough information to understand the concept being coded for and how it should be coded.  However, in practice, researchers often discuss and clarify coding definitions with each other as part of the codebook development or coding process.  If this occurs, these informal decisions may not always be reflected in the final codebook.  Furthermore, researchers implicitly make assumptions about a coder’s prior knowledge when deciding how detailed to make code descriptions.  While these assumptions are normally tested during IRR sessions for human coders, they can be tested earlier for LLMs during codebook development.  Examples of codebook revisions used in this study are provided in Appendix \ref{sec:codebooks}. 

\subsubsection{Tests of Reliability between Human Coders and LLM}
Once the prompts and codebooks are developed and the model codes pass validity testing, a small calibration sample is drawn, and both the model and one or more human coders code the sample documents. At a minimum, human-model IRR metrics are calculated at this stage to determine if a human coder and the model reliably agree on how documents should be coded.  This follows standard practice in content analysis, the only difference being that instead of both coders being humans, one is a human and the other is a model.  Additionally, if the documents are double coded by human coders, we can also compare human-human agreement to human-model agreement to see if they are both high, and ideally, equivalent. This approach is popular in the literature \cite{gilardi_chatgpt_2023, tornberg_chatgpt-4_2023, xiao_supporting_2023} and provides an informative human performance benchmark.  For both reliability assessments, model reasons can be used to provide additional context on disagreements.

If after conducting these assessments the research team determines that the model generated codes are inferior to the human coded responses, then they can either perform additional prompt engineering to modify the prompts and codebook for the underperforming categories or consider alternative approaches for coding the underperforming categories.  These alternatives will depend largely on the type of the failure; discussions of some of these are presented in Section \ref{sec:discuss}.  

\subsubsection{Use LLM to create the Final Coded Dataset}
If model generated codes are deemed non-inferior to the human coded responses, then the last step in LACA is to use the LLM for the final coding.  Since the model predictions are much less time and cost prohibitive than human coding (Section \ref{sec:summary}), either all documents can be coded, or a large random sample of documents can be coded to help stabilize the point estimates and confidence intervals.  This step provides the largest burden reduction compared to a traditional content analysis, since most coding can now be provided by the model rather than human coders.

\subsection{Experiment}
To demonstrate this approach, we compare GPT-3.5 \cite{noauthor_introducing_nodate, shen_chatgpt_2023} to human coders across four publicly available data sets.  GPT-3.5 is a class of LLMs derived from the GPT-3 \cite{brown_language_2020} architecture that has undergone instruction fine-tuning \cite{wei2021finetuned} to improve zero-shot learning performance and/or proximal policy optimization \cite{ouyang2022training} to more closely reflects users’ preferences.  The specific GPT-3.5 model used in this study (gpt-3.5-turbo) is optimized for chat and is currently the primary model used in ChatGPT.  To reduce the variability in responses, the temperature hyperparameter was set to 0 for all model runs.

\subsubsection{Prompts}
To use GPT-3.5 for deductive coding, we wrote a series of prompts that instructed the LLM to perform deductive coding given the codebook and a text (Appendix \ref{sec:prompts}). In some cases, the codebook instructions were updated to clarify the code definitions and standardize format.  The prompt design is inspired by techniques such as role prompting \cite{Schulhoff_Learn_Prompting_2022}, where a role for the AI is assigned (e.g., “you are a qualitative coder performing deductive coding”), and chain-of-thought prompting \cite{wei_chain--thought_2023}, where a complex reasoning task is broken down into sub-tasks and the model is asked to perform the sub-tasks serially.  Outside of examples provided in the existing codebooks, additional few-shot examples for in-context learning \cite{dong2022survey} were not included, making the majority of coding tasks zero-shot.  One design choice of note is the difference in how we approached mutually exclusive vs. non-mutually exclusive coding schemes.  For mutually exclusive code sets, we used a prompt that requests a single code and coding explanation. However, for non-mutually exclusive code sets, we used a prompt that iterates across each code individually, to provide a code and coding explanation for each code in the set separately.

\subsubsection{Data}
Four publicly available data sets were used to demonstrate LACA and assess the deductive coding performance of GPT-3.5 (Table \ref{tab:tab1}). Data sets were considered for inclusion if they contained the original text, coded data, and a codebook, though in some instances, a rudimentary codebook was derived from the category names. The original and revised codebook text for each dataset are available in Appendix \ref{sec:codebooks}.

The \textbf{Trump Tweets} data set consists of a collection of tweets posted by Donald Trump, obtained from the Trump Twitter Archive \cite{brown_search_nodate}, spanning the first two years of his U.S. presidential term. To create the data set, Coe et al. \cite{coe2020marginalized} drew a stratified sample by month, resulting in a set of 2,082 tweets which were then subsequently coded. The codebook includes 13 non-mutually exclusive categories, encompassing both thematic and syntactic attributes. For instance, the "CRIT" category denotes the criticism of another person or idea, while "HSTG" category indicates the usage of hashtags (i.e., words or acronyms beginning with the “\#” character).

The \textbf{Contrarian Claims} data set consists of 2,904 paragraph-length texts from traditional media and social media that make claims about climate change that are contrary to scientific consensus. Coan et al. \cite{coan_computer-assisted_2021} first developed their taxonomy of super-, sub-, and sub-sub-claims by referring to the list of climate myths at skepticalscience.com \cite{noauthor_global_nodate} and categorizing them into main categories. Climate-literate volunteers underwent training before independently coding the documents and were instructed to assign one code per document down to the sub-claim level. The codebook contains 27 sub-claims, as well as a code designated for “None of the above”.

The \textbf{BBC News} data set \cite{greene_practical_2006} consists of 2,225 news articles published by the British Broadcasting Corporation. These articles cover stories in five topic areas from 2004-2005, which serve as the codes in the codebook: technology, business, politics, entertainment, and sports. The coding task aimed to assign each text to a single category. 

The \textbf{Ukraine Water Problems} data set \cite{afanasyev_river_2013} consists of reports addressing water quality issues in Ukraine. A portion of the texts was originally written in Ukrainian and subsequently translated into English, resulting in some instances of imprecision and ambiguity. The codebook consists of 5 non-mutually exclusive categories: environmental problems, pollution, treatment plants or technologies, climatic indicators, and biomonitoring in water or a river basin.

\begin{table}[H]
\caption{Data Set Summary}
\vspace*{2.5mm}
\begin{tabular}{@{}lp{2cm}lccp{4.5cm}cl@{}}
\toprule
\multirow{2}{*}{\textbf{Data Set}} & \textbf{Document}      & \textbf{Mutually}  & \textbf{Codes} & \textbf{Documents} & \multirow{2}{*}{\textbf{Notes}}                                \\
                          & \textbf{Type}          & \textbf{Exclusive} & \textbf{(N)}   & \textbf{(N)}       &                                                       \\ \midrule
Trump Tweets              & Tweets        & No        & 13    & 2,083     & Codebook written informally with short descriptions   \\ \\
\begin{tabular}[c]{@{}l@{}}Contrarian Claims\tablefootnote{We only include observations from the test set of the Contrarian Claims dataset, since the authors state that it went through the most validation.  The full dataset consists of 28,945 observations.}\end{tabular} &
  Blog Posts &
  Yes &
  28 &
  2,904 &
  Mutually exclusive, hierarchical code set.  Codes nuanced   and may have definitions with conceptual overlap. \\ \\
BBC News                  & News Articles & Yes       & 5     & 2,225     & No formal codebook, only class names (e.g., business) \\ \\
\begin{tabular}[c]{@{}l@{}}Ukraine Water Problems\end{tabular} &
  Water Quality Reports &
  No &
  5 &
  100 &
  Brief codebook, but technically complex classes \\ \bottomrule
\end{tabular}
\label{tab:tab1}
\end{table}

\subsubsection{Evaluation}
\label{sec:eval}
We performed evaluations at two levels of granularity: (1) an in-depth case study demonstrating the formative steps of LACA  on the Trump Tweets data set (Section \ref{sec:case-study}), and (2) summary benchmark results across all four data sets comparing the coding performance of LLMs and humans (Section \ref{sec:summary}).

The case study results are divided into two subsections to simulate the two steps of LACA requiring researcher engagement and feedback. The first is an evaluation on a random sample of uncoded documents designed to mimic co-developing a codebook with an LLM ("development set").  The second is an evaluation on the same documents, but now with two sets of human codes, designed to mimic comparing the LLM and human codes for reliability ("calibration set"). For brevity, we only report coding reasons and other qualitative evidence as part of the case study.  Detailed information on the development set evaluations (Section \ref{sec:dev-method}) and calibration set evaluations (Section \ref{sec:cal-method}) are provided below.

The summary benchmark results report similar information as the case study, but in less detail to help highlight trends across the four data sets.  However, detailed quantitative results for all data sets are available in Appendix \ref{sec:detailed_results}. One additional evaluation performed only in the summary benchmark results is a comparison of the wall time taken for the human coders and LLM to code documents. More information on the comparisons performed in the summary benchmark results are provided in Section \ref{sec:bench-method}.

\paragraph{Development Set}
\label{sec:dev-method}
The development set is composed of 100 randomly drawn documents from the Trump Tweets data set and mimics using uncoded documents to develop and test the codebook.  For this set, we produce model generated codes (with coding reasons) and conduct hypothesis tests of randomness.

The tests of randomness for the case study use a two-sided binomial test with a probability of success of 0.5 ($\alpha$ = 0.05). The model reasons are assessed qualitatively, focusing primarily on examples where the model disagrees with the human coders.  We present examples of both model reasons that are compelling, as well as examples of hallucinations (statements posed as facts by the model that cannot be verified from the source material \cite{bang2023multitask}). We also use the model generated reasons to provide additional insight into our quantitative results.

\paragraph{Calibration Set}
\label{sec:cal-method}
The calibration set uses the same 100 observations as the development set but now with two sets of human codes.  One set is the codes reported in the original studies ("original coding") and the other is an independently coded set by our study team ("replicated coding").  This is to help mimic the process of double coding and to provide a more useful human benchmark to compare against the LLM generated codes.  IRR metrics are calculated between the originally published and replicated human codes and between the originally published and LLM generated codes.  

Gwet’s AC1 \cite{gwet2008computing} is our preferred IRR metric because many of the codes are rare and occur infrequently in our calibration sample.  Other popular metrics, such as Cohen’s kappa \cite{cohen1960coefficient}, can suffer in these settings from the “high agreement, low reliability” paradox \cite{falotico2015fleiss, feng2015mistakes, zhao2013assumptions}, where there is little disagreement overall in how the observations are coded but the few disagreements that do occur create large disparities in IRR values.  

The definition of Gwet's AC1 is as follows:

\begin{equation}
Gwet's\:AC1 = {\frac {p_a-p_{e\gamma}}{1-p_{e\gamma}}}
\end{equation}

where $p_a$ is the overall percent agreement and $p_{e\gamma}$ is the chance agreement probability.  Their computation formulas are defined as:

\begin{equation}
p_a = \frac{1}{n} \sum_{i=1}^n \sum_{q=1}^Q \frac{r_{iq}(r_{iq}-1)}{r(r-1)}
\end{equation}

\begin{equation}
p_{e\gamma} = \frac{1}{Q-1} \sum_{q=1}^Q \pi_{q}(1-\pi_{q})
\end{equation}

\begin{equation}
\pi_{q} = \frac{1}{n} \sum_{i=1}^n \frac{r_{iq}}{r}
\end{equation}

where:
\begin{conditions}
 i     &  number of documents rated \\
 q     &  number of categories \\   
 r_{iq} &  number of raters who classified the $i^{th}$ document into the $q^{th}$ category \\
 r &  total number of raters \\
 \pi_{p} & probability that a rater classifies a document into category $q$
\end{conditions}

\paragraph{Summary Benchmark}
\label{sec:bench-method}

The summary benchmark results contain both tests of randomness and IRR metrics across all four data sets. Like the case study, we generated a random sample of 100 documents for each data set and compare the human codes from the published data to (1) the LLM generated codes and ("human-model agreement") (2) those coded by our study team ("human-human agreement").

We use two different hypothesis tests of randomness for the summary benchmark results, depending on the structure of the coding task.  For binary outcomes (Trump Tweets and Ukraine Water Problems), we use a two-sided binomial test with a probability of success of 0.5, and for multiple categorical outcomes (BBC News and Contrarian Claims), we use a chi-squared test against a reference multinomial distribution of equal probabilities. For all assessments, we use an alpha level of 0.05. For the IRR metrics, we again use Gwet's AC1.

Lastly, we calculate the amount of time it took for our study team to code the samples and compare it to the time it takes for GPT-3.5 to generate codes for the same documents.

\section{Results}
\label{sec:results}

\subsection{Case Study: Trump Tweets}
\label{sec:case-study}

\subsubsection{Tests of Randomness (Development Set)}
\label{sec:tests-random}

Table \ref{tab:tab2} reports the results of the binomial tests of randomness for each of the 13 codes in the Trump Tweets data set.  All tests were significant with four exceptions: HSTG (stat: 0.57, p-val: 0.193), ATSN (stat: 0.51, p-val: 0.920), CAPT (stat: 0.48, p-val: 0.764), and INDV (stat: 0.57, p-val: 0.193).  This suggests that for these four categories, the model may be randomly guessing and does not understand the task well enough to produce valid codes.  Three of these codes relate to syntax and formatting of tweet text, including usage of hashtags “\#” (HSTG), at signs “@” (ATSN), or presence of words that only contain capital letters (CAPT).  The INDV code relates to reference of an individual person in a tweet, either named or unnamed.

\begin{table}[H]
\caption{Tests of Randomness for Trump Tweets data set}
\vspace*{2.5mm}
\begin{tabular}{@{}p{2cm}p{8cm}C{2cm}C{2cm}lcc@{}}
\toprule
\textbf{Code} & \textbf{Abbreviated Code Description\tablefootnote{See Table \ref{tbl:trump-code-desc} in Appendix \ref{sec:codebooks} for detailed code descriptions and codebook.}} & \textbf{Estimated Proportion\tablefootnote{The "estimated proportion" is the proportion of the development set that received this code by the LLM.}} & \textbf{P-value} \\ \midrule
HSTG     & Use of hashtag                     & 0.57           & 0.19             \\[1mm] 
ATSN     & Use of @                           & 0.51           & 0.92             \\[1mm] 
CAPT     & Contains words with only capital letters             & 0.48           & 0.76             \\[1mm] 
INDV     & References an individual           & 0.57           & 0.19             \\[1mm] 
CRIT     & Criticizes another person/idea    & 0.30           & 0.00             \\[1mm] 
MEDI     & Makes derogatory statements about news media  & 0.06           & 0.00             \\[1mm] 
FAMY     & References members of Donald Trump's immediate family  & 0.03           & 0.00             \\[1mm] 
PLCE     & References the police              & 0.04           & 0.00             \\[1mm] 
MAGA     & References “Make America Great Again” slogan & 0.06           & 0.00             \\[1mm] 
MARG     & References marginalized group(s)   & 0.02           & 0.00             \\[1mm] 
INTN     & References other countries/leaders & 0.15           & 0.00             \\[1mm] 
PRTY     & References partisan/ideological labels  & 0.23           & 0.00             \\[1mm] 
IMMG     & References immigration             & 0.08           & 0.00             \\ \bottomrule
\end{tabular}
\label{tab:tab2}
\end{table}

\subsubsection{Reasons for Coding Decisions (Development Set)}
\label{code-reasons}

To provide additional context and assess face validity, Table \ref{tab:tab3} provides examples of the model’s coding decisions for exemplar codes that either passed or failed the tests of randomness.  For codes in this table that passed the tests of randomness (MAGA, INTN), the reasons provided largely follow the codebook definitions. In the MAGA example, the model recognizes the Trump campaign’s 2016 presidential slogan (“Make America Great Again”), whereas in the INTN example, the model identifies references to international leaders outside of the US (Qatar) and that the tweet discusses an international issue (funding of radical ideology).  For codes in this table that failed the tests of randomness (HSTG, CAPT), the LLM provided reasons contradict the code definitions and/or contain hallucinations.  In the HSTG example, the model hallucinates that the word “Lowest” contains a hashtag.  In the CAPT example, the model does not recognize that several words in the tweet (“I LOVE YOU ALL”) are all fully capitalized.

\begin{table}[!htb]
\caption{Example LLM Coding Reasons on the Trump Tweets dataset for codes that passed and failed the Tests of Randomness}
\vspace*{2.5mm}
\begin{tabular}{@{}lcp{4.5cm}p{1cm}p{5.5cm}lcl@{}}
\toprule
\textbf{Code} &
  \textbf{Passed Test} &
  \multirow{2}{*}{\textbf{Text}} &
  \textbf{LLM} &
  \multirow{2}{*}{\textbf{LLM Coding Reason}} \\
\textbf{Name} &
  \multicolumn{1}{l}{\textbf{of Randomness}} &
   &
  \multicolumn{1}{l}{\textbf{Coding}} &
   \\ \midrule
MAGA &
  Yes &
  Honored to have Republican Congressional Leadership join me at the @WhiteHouse this evening. Lots to discuss as we continue MAKING AMERICA GREAT AGAIN! https://t.co/b9z5Nfdkcl &
  Yes &
  The tweet explicitly mentions "MAKING AMERICA GREAT AGAIN", which is a direct reference to Donald Trump's 2016 campaign slogan. Therefore, the most applicable code is "Yes". \\ \\
INTN &
  Yes &
  During   my recent trip to the Middle East I stated that there can no longer be funding of Radical Ideology. Leaders pointed to Qatar - look! &
  Yes &
  The   tweet references a country outside of the 50 U.S. states (Qatar) and   discusses an international issue (funding of Radical Ideology). Therefore, it   falls under the code for referencing international topics outside of the USA. \\ \\
HSTG &
  No &
  So   true Wayne and Lowest black unemployment in history! https://t.co/gDxxJdZQUm &
  Yes &
  The   tweet includes a hashtag (\#Lowest) which is not a hyperlink, indicating the   use of a hashtag. Therefore, the code "Yes" is applicable. \\ \\
CAPT &
  No &
  So   sorry people wanting to get into the already packed arena - I LOVE YOU ALL! https://t.co/PFvXrsvgkA &
  No &
  This   tweet does not contain any words in all capital letters, and any   capitalization used is not for emphasis. Additionally, there are no hashtags   or links present in the tweet. Therefore, the code for words in all capital   letters does not apply to this tweet. \\ \bottomrule
\end{tabular}
\label{tab:tab3}
\end{table}

\pagebreak
\subsubsection{Inter-rater Reliability (Calibration Set)}
Table \ref{tab:tab4} outlines the IRR results for the Trump Tweets calibration set between (1) the originally coded data and the LLM coded data (“human-model agreement”) and (2) the originally coded data and our study teams coding (“human-human agreement”).  Gwet’s AC1 for human-model agreement is generally high across codes (>0.76), with the exception of codes that failed the hypothesis tests of randomness (HSTG: 0.18; ATSN: 0.58; CAPT: 0.36; INDV: 0.50).  Gwet’s AC1 for the human-human agreement is also generally high (>0.73) and comparable to the human-model agreement for all codes, except again for those that failed the test of randomness, in which the human-human agreement is far higher (HSTG: 0.96; ATSN: 1.00; CAPT: 0.93; INDV: 0.79) than the human-human agreement.  

\begin{table}[!htb]
\caption{Human-Human and Human-Model Agreement on the Trump Tweets Data Set}
\vspace*{2.5mm}
\begin{tabular}{@{}p{2.5cm}C{4.5cm}C{4.5cm}pcc@{}}
\toprule
\multirow{2}{*}{\textbf{Code}} & \multicolumn{2}{c}{\textbf{Gwet's AC1}}              \\ \cmidrule(l){2-3} 
                               & \textbf{Human-Human Agreement} & \textbf{Human-Model Agreement} \\ \midrule
HSTG                           & 0.96                         & 0.18                  \\[0.5mm]
ATSN                           & 1.00                         & 0.58                  \\[0.5mm]
CRIT                           & 0.73                         & 0.76                  \\[0.5mm]
MEDI                           & 1.00                         & 0.96                  \\[0.5mm]
FAMY                           & 0.97                         & 0.96                  \\[0.5mm]
PLCE                           & 1.00                         & 0.98                  \\[0.5mm]
MAGA                           & 0.99                         & 0.98                  \\[0.5mm]
CAPT                           & 0.93                         & 0.36                  \\[0.5mm]
INDV                           & 0.79                         & 0.5                   \\[0.5mm]
MARG                           & 0.97                         & 0.94                  \\[0.5mm]
INTN                           & 0.86                         & 0.81                  \\[0.5mm]
PRTY                           & 0.81                         & 0.76                  \\[0.5mm]
IMMG                           & 0.99                         & 0.97                  \\ \bottomrule
\end{tabular}
\label{tab:tab4}
\end{table}

\pagebreak
\subsubsection{Reasons for Coding Decisions (Calibration Set)}
\label{code-reasons-cal}
Coding reasons can also be used in the calibration phase to further refine the codebook and identify areas of disagreement between models and human coders.  Table \ref{tab:tab5} provides model coding reasoning examples for which the human coders agree but the reasons indicate that the model does not understand the codebook and/or text.  The first two rows (FAMY, INDV) showcase the model being too liberal in its interpretation of the codebook.  In the FAMY example (a code for if the tweet mentions any of Donald Trump’s immediate family members), the model codes the tweet as being relevant because it references the U.S. Secretary of Homeland Security, Kirstjen Nielsen, a member of his “political family.”  In the INDV example (a code for if the tweet mentions an individual, named or unnamed), the model codes the tweet as being relevant because it interprets the Democratic Party as being composed of “a group of individuals.”  

In contrast, the last two rows (PLCE, IMMG) showcase the model being too strict in its interpretation of the codebook.  In the PLCE example (a code for if the tweet mentions the police), the model codes the tweet as not being relevant because it references the Florida Highway Patrol Troopers, which it reasons “is a specific branch of law enforcement and not a general reference to police.”  In the IMMG example (a code for if the tweet references U.S. immigration), the model codes the tweet as not being applicable because it reasons, “border security… does not necessarily equate to immigration,” even though immigration was a justification for border security measures during Trump’s 2016 campaign, such as building a wall on the U.S. southern border \cite{gravelle_politics_2018}.

\begin{table}[!htb]
\caption{Example LLM Coding Reason on the Trump Tweets data set for which the model too loosely (FAMY, INDV) or strictly (PLCE, IMMG) interprets the codebook.}
\vspace*{2.5mm}
\begin{tabular}{@{}lp{4cm}cccp{4.5cm}l@{}}
\toprule
\textbf{Code Name} &
  \textbf{Text} &
  \textbf{\begin{tabular}[c]{@{}c@{}}Original\\    Coding\end{tabular}} &
  \textbf{\begin{tabular}[c]{@{}c@{}}Replicated\\    Coding\end{tabular}} &
  \textbf{\begin{tabular}[c]{@{}c@{}}LLM \\    Coding\end{tabular}} &
  \textbf{LLM Coding Reason} \\ \midrule
FAMY &
  Homeland Security @SecNielsen did a fabulous job yesterday at the press conference explaining security at the border and for our country while at the same time recommending changes to obsolete \& nasty laws which force family separation. We want “heart” and security in America! &
  No &
  No &
  Yes &
  In this tweet, Donald Trump references Kirstjen Nielsen, who is his Secretary of Homeland Security. This counts as a family reference, as Nielsen is a member of his administration and therefore considered part of his "political family". \\ \\
INDV &
  Never seen the Republican Party so unified. No “Cave”  on the issue of Border and National Security. A beautiful thing to see especially when you hear the new rhetoric spewing from the mouths of the Democrats who talk Open Border High Taxes and Crime. Stop Criminals \& Drugs now! &
  No &
  No &
  Yes &
  The tweet references the Democratic Party, which is a group of individuals, and also mentions "new rhetoric spewing from the mouths of the Democrats", indicating that specific individuals within the party are being referenced. Additionally, the tweet mentions "Criminals \& Drugs", which could be interpreted as a reference to specific individuals involved in criminal activity. Therefore, the tweet references individual people and should be coded as a "Yes". \\ \\ 
PLCE &
  Florida Highway Patrol Troopers are all en route to the Panhandle from all across the state of Florida - to help those affected by \#HurricaneMichael. If you see them be sure to shake their hands and say THANK YOU! \#LESM https://t.co/rB7uNBudY5 &
  Yes &
  Yes &
  No &
  The tweet does not reference the police or individual police officers. It mentions Florida Highway Patrol Troopers, but this is a specific branch of law enforcement and not a general reference to police. Therefore, the "Does this tweet reference the police?" code is not applicable. \\ \\ 
IMMG &
  I am in the White House working hard. News reports concerning the Shutdown and Syria are mostly FAKE. We are negotiating with the Democrats on desperately needed Border Security (Gangs Drugs Human Trafficking \& more) but it could be a long stay. On Syria we were originally... &
  Yes &
  Yes &
  No &
  The tweet does not reference immigration or any related topics such as the wall, southern border, ICE, separation of families, or the travel ban. The focus is on negotiations with Democrats regarding border security, but this does not necessarily equate to immigration. Therefore, the most applicable code is "No." \\ \bottomrule
\end{tabular}
\label{tab:tab5}
\end{table}

However, LLM coding reasons can also help identify areas of misunderstanding for human coders. Table \ref{tab:tab6} provides examples of model coding reasons for which the model and one human coder agrees, but the other does not (PRTY, FAMY), as well as an example where both human coders disagree with the model, but the model reasoning seems valid (IMMG).  In the PRTY example (a code for if the tweet mentions a U.S. political party), the model and the study team coder agree the code should apply to the tweet because it mentions the DNC, an acronym for the Democratic National Committee.  In the FAMY example, the original code and model agree that the code should apply because it references Melania Trump, Donald Trump’s wife.  In the final IMMG example, although neither human coder believes the code should apply, the model provides a compelling argument for its inclusion by stating that the tweet includes the DACA (Deferred Action for Childhood Arrivals) program, which is a U.S. immigration policy \cite{johnson_lessons_2018}.

\begin{table}[!htb]
\caption{Example LLM Coding Reasons on the Trump Tweets data set for which the model produces valid \\
code reasons but disagrees with one human coder (PRTY, FAMY) or both human coders (IMMG). }
\vspace*{2.5mm}
\begin{tabular}{@{}lp{4cm}cccp{4.5cm}l@{}}
\toprule
\textbf{Code Name} &
  \textbf{Text} &
  \textbf{\begin{tabular}[c]{@{}c@{}}Original\\    Coding\end{tabular}} &
  \textbf{\begin{tabular}[c]{@{}c@{}}Replicated\\    Coding\end{tabular}} &
  \textbf{\begin{tabular}[c]{@{}c@{}}LLM \\    Coding\end{tabular}} &
  \textbf{LLM Coding Reason} \\ \midrule
PRTY &
  Everyone here is talking about why John Podesta refused to give   the DNC server to the FBI and the CIA. Disgraceful! &
  No &
  Yes &
  Yes &
  The   tweet references the DNC (Democratic National Committee) which is a political   party in the US. Therefore, it falls under the "References to Democrat   or Liberal" category in the codebook and should be coded as   "Yes". \\ \\
FAMY &
  President @EmmanuelMacronThank you for inviting Melania and   myself to such a historic celebration in France. \#BastilleDay \#14juillet   https://t.co/KZjJQOo0H1 &
  Yes &
  No &
  Yes &
  The tweet includes a reference to Melania, who is a member of   Donald Trump's immediate family. Therefore, the code "Yes" is   applicable. \\ \\ 
IMMG &
  Federal Judge in Maryland has just ruled that‚ “President Trump   has the right to end DACA”. President Obama had 8 years to fix this problem   and didn’t. I am waiting for the Dems they are running for the hills! &
  No &
  No &
  Yes &
  This tweet references DACA, which is a program related to   immigration in the US. Additionally, Trump mentions the Democrats   "running for the hills," which suggests that this tweet is related   to a political debate around immigration policy. \\ \bottomrule
\end{tabular}
\label{tab:tab6}
\end{table}

\pagebreak
\subsection{Summary Benchmark Results}
\label{sec:summary}

Table \ref{tab:tab7} summarizes the IRR results and tests of randomness across all four data sets.  The tests of randomness pass for all codes except for those previously identified in the Trump Tweets data set (HSTG, ATSN, CAPT, INDV) and for two codes in the Ukraine Water Problems data set (env problems, pollution). After reviewing the model reasoning for the failing codes in the Ukraine Water Problems data set, the model reasons seem largely valid and appear to be coding for some version of the concepts “environmental problems” and “pollution.”  This suggests that the tests are failing solely because the prevalence of the model’s coding decisions are close to 50\% and not because they are randomly guessing.  Interestingly, the two codes in the Ukraine Water Problems data set that failed were still among the lowest human-human agreement and human-model agreement across data sets.








Comparatively, the human-model agreement is higher than the human-human agreement for the 3 of the 5 codes in the Ukraine Water Problems data set and across codes for the BBC News data set.  The human-model agreement is lower than the human-human agreement for 2 of the 5 codes in the Ukraine Water Problems data set and across codes for the Contrarian Claims data set.   Appendix \ref{sec:detailed_results} contains detailed results and comparisons for all data sets. 

\begin{table}[!htb]
\caption{Summary Benchmark Results Across Data Sets}
\vspace*{2.5mm}
\begin{tabular}{@{}llccc@{}}
\toprule
\multirow{2}{*}{\textbf{Dataset}} & \multirow{2}{*}{\textbf{Code}} & \multicolumn{2}{c}{\textbf{Gwet's AC1}} & \multicolumn{1}{c}{\multirow{2}{*}{\textbf{Tests of Randomness}}} \\ \cmidrule(lr){3-4}
 &   & \textbf{Human-Human}  & \textbf{Human-Model}  &  \multicolumn{1}{c}{\textbf{(p-value)}}                         \\ \midrule
Trump Tweets      & HSTG          & 0.96 & 0.18 & 0.19 \\[0.5mm] 
Trump Tweets      & ATSN          & 1.00 & 0.58 & 0.92 \\[0.5mm] 
Trump Tweets      & CRIT          & 0.73 & 0.76 & 0.00 \\[0.5mm] 
Trump Tweets      & MEDI          & 1.00 & 0.96 & 0.00 \\[0.5mm] 
Trump Tweets      & FAMY          & 0.97 & 0.96 & 0.00 \\[0.5mm] 
Trump Tweets      & PLCE          & 1.00 & 0.98 & 0.00 \\[0.5mm] 
Trump Tweets      & MAGA          & 0.99 & 0.98 & 0.00 \\[0.5mm] 
Trump Tweets      & CAPT          & 0.93 & 0.36 & 0.76 \\[0.5mm] 
Trump Tweets      & INDV          & 0.79 & 0.50 & 0.19 \\[0.5mm] 
Trump Tweets      & MARG          & 0.97 & 0.94 & 0.00 \\[0.5mm] 
Trump Tweets      & INTN          & 0.86 & 0.81 & 0.00 \\[0.5mm] 
Trump Tweets      & PRTY          & 0.81 & 0.76 & 0.00 \\[0.5mm] 
Trump Tweets      & IMMG          & 0.99 & 0.97 & 0.00 \\[0.5mm] 
Ukraine Water     & env\_problems & 0.23 & 0.64 & 0.62 \\[0.5mm] 
Ukraine Water     & pollution     & 0.59 & 0.55 & 0.62 \\[0.5mm] 
Ukraine Water     & treatment     & 0.84 & 0.88 & 0.00 \\[0.5mm] 
Ukraine Water     & climate       & 0.97 & 0.87 & 0.00 \\[0.5mm] 
Ukraine Water     & biomonitoring & 0.51 & 0.86 & 0.00 \\[0.5mm] 
BBC News          & All           & 0.76 & 0.85 & 0.00 \\[0.5mm] 
Contrarian Claims & All           & 0.65 & 0.59 & 0.00 \\ \bottomrule
\end{tabular}
\label{tab:tab7}
\end{table}

Table \ref{tab:tab8} depicts a comparison of the coding time per observation for the human coders and GPT-3.5 to perform the deductive coding, defined as the number of seconds per document taken to provide a code.  Note that GPT-3.5 also produces model reasons for each coding decision as part of its output, which differs from the human coders that only provide a code.  Additionally, requests to the OpenAI API were made serially in this experiment, making this assessment more conservative than if optimized to run in parallel.

\begin{table}[htb!]
\caption{Coding Time per Document}
\vspace*{2.5mm}
\begin{tabular}{@{}lcc@{}}
\toprule
\multirow{2}{*}{\textbf{Dataset}} & \multicolumn{2}{c}{\textbf{Coding Time (seconds / document)}} \\ \cmidrule(l){2-3} 
                         & \textbf{Human Coder}                & \textbf{LLM Coder}               \\ \midrule
Trump Tweets             & 72                         & 52                      \\[0.5mm]
Ukraine Water            & 108                        & 16                      \\[0.5mm]
BBC News                 & 72                         & 4                       \\[0.5mm]
Contrarian Claims        & 144                        & 4                       \\ \bottomrule
\end{tabular}
\label{tab:tab8}
\end{table}

Across data sets, human coders took longer than GPT-3.5 to provide codes, though there are some relative differences worth noting.  The largest discrepancy in coding time comes from the Contrarian Claims data set (humans: 144 sec/doc; LLM: 4 sec/doc). This data set has longer texts and far more code categories than the others, making it the most burdensome coding task for the human coders.  This difference is exacerbated due to its mutually exclusive measures that only requires a single request to GPT-3.5 to return a coding decision and reason.  Conversely, the Trump Tweets data set has the closest coding times between human coders (72 sec/doc) and model coders (52 sec/doc).  This is because tweets are short and the coding task is less nuanced, making it easier for human coders to review.  GPT-3.5 takes relatively longer on this task because we made separate requests for each of the 13 codes to ensure that the model focused on each code individually and would be able to produce a model explanation separately for each code. This design choice was made because the codes in this task are not mutually exclusive and can each apply to any given tweet.

\section{Discussion}
\label{sec:discuss}
In this work we present LACA, a holistic approach to integrate LLMs into deductive coding.  We demonstrate this method with an in-depth case study and an empirical evaluation across four publicly available data sets, comparing both coding results and coding time between humans and a commercial LLM. 

Across data sets, the tests of randomness helped identify codes for which there were substantial human-model disagreement, suggesting that it can help initially assess when the model does not understand the measure without requiring human coded data.  However, as seen in the Ukraine Water Problems data set, the tests may also fail to reject the null hypothesis if the model is performing well but the true base rate is close to an equal probability. For this reason, pairing the tests with manual assessment of the model reasons is crucial to properly interpreting the results. Though others have assessed LLMs by comparing their output to random chance \cite{ruis_large_2022,wei_emergent_2022}, to our knowledge, this is the first work to formalize this comparison with hypothesis testing.  

By analyzing the codes that failed the tests of randomness in the Trump Tweets case study (HSTG, ATSN, CAPT, INDV), we observe two types of issues that may result in poor model coding performance.  The set of “formatting codes” (HSTG, ATSN, CAPT) likely fail due to complications introduced by GPT-3.5’s byte-pair encoding (BPE) tokenizer \cite{brown_language_2020}. Instead of segmenting text into words, BPE tokenization segments text into words, word fragments, or characters chosen to minimize the encoding length on the original model’s corpus.  While common smaller words may have their own BPEs, longer or rare words may be encoded with several BPEs, with completely novel words potentially being encoded as combinations of individual letter or character BPEs.  Because models using the GPT architecture do not default to using characters representations, it can be challenging for them to learn character-level aspects of language, such as word length \cite{efrat_lmentry_2022} and presence of certain characters \cite{rogers_whats_nodate}. These types of model issues are inherent to the LLM and cannot be fully addressed with additional prompt engineering; understanding how the underlying LLM model being used for LACA is designed and trained is important for assessing these limitations.  However, character-level aspects of language are often of limited value to social scientists and can be identified if needed using simpler methods with high accuracy (e.g., regular expressions).  Unlike the formatting codes, the INDV code seems to fail in several instances because the model assigns any mention of a group to this code, since it reasons that a group consists of several individuals.  We consider this a codebook issue that has a higher chance of being addressed through codebook clarification and prompt engineering.

Qualitatively, we found that producing model reasons for coding decisions helps human coders assess model performance, prompt quality, and build trust in predictions.  Examples in the Results section demonstrate model generated reasons that helped in identifying hallucinations and reasoning errors and suggests ways in which the codebook can be modified to improve performance.  Additionally, we found that model generated reasons can help human coders reflect on their own coding decisions, which can in turn inform revisions to measure definitions.  Anecdotally, we find that the process of making the codebook more explicit for the model also tends to help improve the instruction readability and comprehension for human coders.  This finding may be less robust if using LLMs that have not gone through the same amount of fine tuning that the GPT-3.5 series of models received, where prompt engineering that more closely mimics next token prediction may be required \cite{liu2023pre}.

Across all data sets, the IRR results suggest that for many cases, GPT-3.5 codes at a level of agreement comparable to human coders.  In cases where the model far underperformed human coders, these usually could be detected early with the hypothesis tests of randomness.  This finding largely agrees with prior research on LLMs ability to generate annotation for NLP tasks, which suggests that GPT-3.5 or GPT-4 meet or exceed the performance of crowd workers \cite{gilardi_chatgpt_2023}, and in some instances, also experts \cite{tornberg_chatgpt-4_2023}.  Interestingly, our results show more promising LLM results than the prior work using LLM for deductive coding \cite{xiao_supporting_2023}, though it’s challenging to draw strong conclusions based on differences in prompting strategies, data sets, metrics, and qualitative coding tasks. One hypothesis for the seemingly larger gap in performance between humans and LLMs in \cite{xiao_supporting_2023} is that their human-human agreement was based on double-coding of multiple experts who developed the original codebook.  This contrasts with our replication focused human-human agreement, which compares the coding in the original studies to an independent coding using solely the published codebooks to derive measure definitions and meanings. Additionally, our coders have no specialized knowledge or expertise in the domains of the coding tasks used in this study, which likely accounts for some of the lower human-human IRR values observed in the more technical data sets (Contrarian Claims and Ukraine Water Problems). We hope that by using publicly available data sets, future research can build on this work and help improve the use of LLMs for deductive coding.

Overall, we believe our findings have several implications for the future practice of deductive coding and related research methods.  First, based on the reduced coding time and, in many cases, comparable results to human coders, we believe LACA and LLMs more generally can be an effective method for deductive coding.
However, despite these implications, we do not consider LACA as a replacement for qualitative researchers, but rather, a tool to support accelerating the latter stages of deductive coding that tend to be more manually taxing and less fulfilling for researchers.  Second, the adoption of LLMs in deductive coding will likely require different types of reporting and documentation to allow for effective reproducibility and critique.  In addition to codebooks and reporting of IRR metrics, researchers using LLMs for deductive coding should also include prompts and model details whenever possible.  Assuming the text itself is also not sensitive in nature, providing the document text, any human or model generated codes, and model generated reasons will also help other researchers assess the validity and reliability of model results.  This is particularly important due to the stochastic nature of generative LLMs that can make it challenging to reproduce the exact same model output in future runs.  To promote this reporting standard, we provide the prompts, codebooks, and model details in the Methods and Appendix and make the coded data sets and model reasons for the calibration set publicly available online \cite{chew_llm-assisted_nodate}.  Lastly, though this work focuses on deductive coding, we believe the approach is also informative for the burgeoning LLM for annotation literature \cite{ding_is_2022, gilardi_chatgpt_2023, tornberg_chatgpt-4_2023, zhu_can_2023}.  In particular, we anticipate the components of LACA that do not require prior labeled data (tests of randomness and model generated reasons) to be especially useful for researchers aiming to generate LLM labeled data.

Our work presents several limitations and opportunities for future work.  First, to match the output in the original data sets, we forced the model to choose either Yes/No or select a single code from a set of mutually exclusive code categories.  However, occasionally the model would return results indicating that the coding tasks were not applicable or that there was not enough information to code the document.  This phenomenon has been documented in prior work using the GPT-3.5 series of models \cite{zhu_can_2023} and is potentially a useful uncertainty quantification mechanism that can better integrated into LACA (e.g., by allowing the model to return “Yes”, “No”, or “I Don’t Know”, along with their reasoning for the decision).  Second, since the objective of this work is to provide a holistic approach for performing deductive coding with LLMs, as opposed to focusing primarily on how well LLMs can perform deductive coding, we did not perform extensive prompt engineering, nor did we assess the coding performance across a wide variety of LLMs. Given the pace that both LLMs and prompt engineering techniques are evolving, we anticipate future work will greatly improve on the empirical LLM performance reported here.  Lastly, a major limitation of using this approach is also one of the intended benefits; researchers using LACA will inevitably end up reading less documents, which in some cases, may limit the development of new theory and discovering themes not proposed by the research team \emph{a priori}.  We anticipate that future work using LLMs to support inductive coding \cite{gao_collabcoder_2023} could be a promising approach to help address these issues. 

\section{Availability of data and materials}
All datasets used in this paper are publicly available and sources are provided in the main manuscript.  Additional coded data is available in \cite{chew_llm-assisted_nodate}.

\section*{Competing Interests}
The authors declare that they have no competing interests.

\section*{Funding}
This work is solely funded by RTI International. 

\section*{Authors’ Contributions}
RC and AK designed the study and secured funding. JB, MW, and RC developed prompts, ran analyses, and provided data management. JS coordinated qualitative coding and coded documents. RC lead writing, with contributions from JB, MW, JS, and AK.  All authors approved the final manuscript.

\section*{Acknowledgments}
We would like to acknowledge Courtney Richardson and Larisa Albers for their coding contributions and Peter Baumgartner for his feedback.

\bibliographystyle{unsrt}  
\bibliography{references}  

\appendix

\pagebreak
\section{Prompts}
\label{sec:prompts}
Figure \ref{fig:figS1} through Figure \ref{fig:figS4} depict the prompts used for each data set. The \{code\} block indicates where the codebook text is entered (see Section \ref{sec:codebooks} for codebooks) and the \{text\} block indicates where the text for each document is entered.  The first "Codebook" and "Tweet"/"Text" references are included to help the model learn the expected input format and the first "Code" reference is to specify the output categories available for the model to use.

\begin{figure}[!htb]
    \caption{Trump Tweets Prompt}
    {\fontfamily{qcr}\selectfont\small\
    \noindent\fbox{%
        \parbox{\textwidth}{%
            You are a qualitative coder who is annotating tweets from Donald Trump's Twitter feed. \\
            
            To code this tweet, do the following: \\
            - First, read the codebook and the tweet. \\
            - Next, decide which code is most applicable and explain your reasoning for the coding decision. \\
            - Finally, print the most applicable code and your reason for the coding decision. \\
            
            Use the following format: \\
            
            Codebook:\\
            \texttt{-{}-{}-}\\
            codebook here\\
            \texttt{-{}-{}-}\\
            
            Tweet:\\
            \texttt{-{}-{}-}\\
            tweet here\\
            \texttt{-{}-{}-}\\
            
            Code:\\
            \texttt{-{}-{}-}\\
            Yes or No\\
            \texttt{-{}-{}-}\\
            
            Codebook:\\
            \texttt{-{}-{}-}\\
            \{code\}\\
            \texttt{-{}-{}-}\\
            
            Tweet:\\
            \texttt{-{}-{}-}\\
            \{text\}\\
            \texttt{-{}-{}-}\\
            
            Code:\\
        }%
    }
    }
    \label{fig:figS1}
\end{figure}

\begin{figure}[!htb]
    \caption{BBC News Prompt}
    {\fontfamily{qcr}\selectfont\small\
    \noindent\fbox{%
        \parbox{\textwidth}{%
            You are a qualitative coder who is annotating tweets from news stories from the BBC. \\
            
            To code this text, do the following: \\
            - First, read the codebook and the text. \\
            - Next, decide which code is most applicable and explain your reasoning for the coding decision. \\
            - Finally, print the most applicable code and your reason for the coding decision. \\
            
            Use the following format: \\
            
            Codebook:\\
            \texttt{-{}-{}-}\\
            codebook here\\
            \texttt{-{}-{}-}\\
            
            Text:\\
            \texttt{-{}-{}-}\\
            text here\\
            \texttt{-{}-{}-}\\
            
            Code:\\
            \texttt{-{}-{}-}\\
            business, entertainment, politics, sport, or tech\\
            \texttt{-{}-{}-}\\
            
            Codebook:\\
            \texttt{-{}-{}-}\\
            {code}\\
            \texttt{-{}-{}-}\\
            
            Text:\\
            \texttt{-{}-{}-}\\
            {text}\\
            \texttt{-{}-{}-}\\
            
            Code:\\
        }%
    }
    }
    \label{fig:figS2}
\end{figure}

\begin{figure}[!htb]
    \caption{Ukraine Water Problems Prompt}
    {\fontfamily{qcr}\selectfont\small\
    \noindent\fbox{%
        \parbox{\textwidth}{%
            You are a qualitative coder who is annotating water quality reports. \\
            
            To code this text, do the following: \\
            - First, read the codebook and the text. \\
            - Next, decide which code is most applicable and explain your reasoning for the coding decision. \\
            - Finally, print the most applicable code and your reason for the coding decision. \\
            
            Use the following format: \\
            
            Codebook:\\
            \texttt{-{}-{}-}\\
            codebook here\\
            \texttt{-{}-{}-}\\
            
            Text:\\
            \texttt{-{}-{}-}\\
            text here\\
            \texttt{-{}-{}-}\\
            
            Code:\\
            \texttt{-{}-{}-}\\
            Yes or No\\
            \texttt{-{}-{}-}\\
            
            Codebook:\\
            \texttt{-{}-{}-}\\
            {code}\\
            \texttt{-{}-{}-}\\
            
            Text:\\
            \texttt{-{}-{}-}\\
            {text}\\
            \texttt{-{}-{}-}\\
            
            Code:\\
        }%
    }
    }
    \label{fig:figS3}
\end{figure}

\begin{figure}[!htb]
    \caption{Contrarian Claims Prompt}
    {\fontfamily{qcr}\selectfont\small\
    \noindent\fbox{%
        \parbox{\textwidth}{%
            You are a qualitative coder who is annotating conservative blog posts related to climate change. \\
            
            To code this text, do the following: \\
            - First, read the codebook and the text. \\
            - Next, decide which code is most applicable and explain your reasoning for the coding decision. \\
            - Finally, print the most applicable code and your reason for the coding decision. \\
            
            Use the following format: \\
            
            Codebook:\\
            \texttt{-{}-{}-}\\
            codebook here\\
            \texttt{-{}-{}-}\\
            
            Text:\\
            \texttt{-{}-{}-}\\
            text here\\
            \texttt{-{}-{}-}\\
            
            Code:\\
            \texttt{-{}-{}-}\\
            1.1, 1.2, 1.3, 1.4, 1.5, 1.6, 1.7, 1.8, 2.1, 2.2, 2.3, 2.4, 2.5, 3.1, 3.2, 3.3, 3.4, 3.5, 3.6, 4.1, 4.2, 4.3, 4.4, 4.5, 5.1, 5.2, 5.3, or 0.0\\
            \texttt{-{}-{}-}\\
            
            Codebook:\\
            \texttt{-{}-{}-}\\
            {code}\\
            \texttt{-{}-{}-}\\
            
            Text:\\
            \texttt{-{}-{}-}\\
            {text}\\
            \texttt{-{}-{}-}\\
            
            Code:\\
        }%
    }
    }
    \label{fig:figS4}
\end{figure}

\clearpage
\section{Codebooks}
\label{sec:codebooks}

\begin{longtable}{@{}lp{6cm}p{7cm}ll@{}}
\caption{Trump Tweets Codebook}\\
\cmidrule(r){1-3}
\textbf{Code Name} &
  \textbf{Original Code Description} &
  \textbf{Revised Code Description} &
   &
   \\* \cmidrule(r){1-3}
\endfirsthead
\endhead
HSTG &
  Hashtag used?  Exclude   links from this decision. &
  Are hashtags used in this tweet?    Exclude hyperlinks from this decision. &
   &
   \\ \\
ATSN &
  @ used?  Include @ present   in retweets.  Exclude links from this   decision. &
  Are at signs ("@") used in this tweet?  Include "@" that are present in   retweets.  Exclude hyperlinks from this   decision. &
   &
   \\ \\
CRIT &
  Criticizes another person/idea (not his own)?  If he suggests at any point in the tweet   that some person or some entity did something wrong, code yes.  ex. “That is not Free or Fair Trade it is   Stupid Trade!” &
  Does Donald Trump criticize another person or idea in this   tweet?  If the author suggests at any   point in the tweet that some person or entity did something wrong, code   'Yes'.  Ex. “That is not Free or Fair   Trade it is Stupid Trade!” &
   &
   \\ \\
MEDI &
  Derogatory/condescending statements about news media?  Individuals journalists talked about in   these terms are a yes.  References to   social media as a general entity are coded “no”.  References to specific news media’s social   media accounts is “yes”. &
  Does Donald Trump make derogatory or condescending statements   about the news media in this tweet?    Include statements made about individual journalists.  References to social media as a general   entity are coded “No”.  References to   specific news media’s social media accounts is “Yes”. &
   &
   \\ \\
FAMY &
  References members of his immediate family?  A family reference to an individual will   also be a yes for INDV.  Do not count   Trump self-references as family; must be someone other than him. &
  Does Donald Trump reference members of his immediate family in   this tweet?  A family reference to an   individual should be coded "Yes".    Do not count Donald Trump's self-references as family; he must   reference someone other than himself. &
   &
   \\ \\
PLCE &
  References the police?  This   category captures only references to police or individual police officers,   not to broader issues of crime, etc. &
  Does this tweet reference the police?  This category captures only references to   police or individual police officers, not to broader issues of crime,   criminal justice, etc. &
   &
   \\ \\
MAGA &
  Reference to this campaign slogan?  Only count forms of the slogan, such as   "MAGA" or "Make America Great Again".  General references to America and greatness   are not enough alone. &
  Does this tweet reference Donald Trump's 2016 campaign   slogan?  Only count forms of the   slogan, such as "MAGA" or "Make America Great   Again".  General references to   America and greatness should be coded as "No". &
   &
   \\ \\
CAPT &
  Capital letters used?  Any   use of capital letters to designate emphasis; at least one full word in caps   must be present for a yes code.    Acronyms do not garner a yes code on their own.  Exclude hashtags and links from this   decision (include hashtags that aren’t acronyms; Ex. \#CHANGETHELAWS). &
  Are there words that contain only capital letters in this   tweet?  Any use of capital letters to   designate emphasis should be considered. At least one full word in caps must   be present.  Acronyms do not garner a "Yes"   code on their own.  Exclude hashtags   and links from this decision (include hashtags that aren’t acronyms; Ex.   \#CHANGETHELAWS). &
   &
   \\ \\
INDV &
  References an individual?    Exclude self-references.  Any   person’s name is a yes (even if within group, such as Obama Administration);   unnamed individuals are a yes, so long as its clear a single person is being   called out.  People tagged with @ count   as a 1, but the @ is not necessary to code 1. &
  Does this tweet reference an individual person?  Exclude self-references to Donald   Trump.  Any person’s name should be   coded as a "Yes". Unnamed individuals should be coded as a   "Yes", so long as it is clear a single person is being   referenced.  People tagged with an @   count as a "Yes", but the @ is not necessary to code a   "Yes". &
   &
   \\ \\
MARG &
  Explicit references to marginalized group(s)?  Women; religious minorities; native   communities; racial/ethnic groups; people with disabilities; sexual/gender   identity.  A single individual is   insufficient to code yes unless the broader group is also mentioned.  Do not code generic references to everyone   that happen to include the group, e.g., “men and women”.  Ex. “Pocahontas” is a yes. &
  Does the tweet explicitly reference a marginalized group or   groups?  Examples of marginalized   groups include women, religious minorities, native communities, racial or   ethnic groups, people with disabilities, and sexual/gender identity.  Mention of a single individual belonging to   one of these groups is insufficient to code "Yes", unless the   broader group is also mentioned.  Do   not code generic references to everyone that happen to include the group   (Ex., “men and women”).  Ex.   “Pocahontas” should be coded as a "Yes". &
   &
   \\ \\
INTN &
  References other countries/leaders?  Anything outside the 50 states.  Include discussion of trade, NATO.  Don’t look up terms you don’t know. &
  Does this tweet reference international topics outside of USA,   such as other countries or international leaders?  Include only references outside the 50 U.S.   states.  Include discussions of trade   and NATO.  Do not look up terms you   don’t know. &
   &
   \\ \\
PRTY &
  Reference to partisan/ideological labels?  0=No label presents;   1=Republican/Conservative; 2=Democrat/Liberal; 3=Both.  Code this strictly, such that only actual   forms of these specific terms garner codes.    “Conservative so-and-so” is a yes.    Examples: “The other side” = 0; “Dems” = 2; “We need more conservative   policies” = 1. &
  Does this tweet reference US political parties?  References to Republican or Conservative is   "Yes". References to Democrat or Liberal is "Yes".  Code this strictly, such that only actual forms   of these specific terms garner codes. Ex. “Conservative so-and-so” is a   "Yes".  Ex. “The other side”   is a "No".  Ex. “Dems” is a   "Yes" 2. Ex. “We need more conservative policies” is a   "Yes". &
   &
   \\ \\
IMMG &
  References immigration?    The wall, the southern border, ICE, separation of families, the travel   ban, etc.  Code a 1 if it is clear he’s referencing immigration, even   without using that word (e.g., “Big increase in traffic into our country from   certain areas while our people are far more vulnerable” would be a 1).  Ex. “ZIKA virus” is a yes. &
  Does this tweet reference immigration in the US?  Include references to the wall, the   southern border, ICE, separation of families, the travel ban, etc. Code   "Yes" if it is clear that Donald Trump is referencing immigration,   even if he does not use the word immigration (Ex., “Big increase in traffic   into our country from certain areas while our people are far more vulnerable”   would be a "Yes").  Ex. “ZIKA   virus” is a "Yes". &
   &
   \\* \cmidrule(r){1-3}
\label{tbl:trump-code-desc}
\end{longtable}

\clearpage
\begin{table}[!htb] 
\caption{Ukraine Water Problems Codebook}
\vspace*{2.5mm}
\begin{tabular}{@{}lp{6cm}p{6cm}ll@{}} 
\cmidrule(r){1-3}
\textbf{Code   Name} & \textbf{Original   Code Description}        & \textbf{Revised   Code Description}         &  &  \\ \cmidrule(r){1-3}
env\_problems        & Is the text about an environmental problem? & Is the text about an environmental problem? &  &  \\ \\
pollution            & Is the text about environmental pollution?  & Is the text about environmental pollution?  &  &  \\ \\
treatment &
  Is the text about treatment plants or environmental   technologies? &
  Is the text about treatment plants or environmental   technologies? &
   &
   \\ \\
climate              & Is the text about climatic indicators?      & Is the text about climatic indicators?      &  &  \\ \\
biomonitoring &
  Is the text about biological, biotic monitoring in water or in a   river basin? &
  Is the text about biological, biotic monitoring in water or in a   river basin? &
   &
   \\ \cmidrule(r){1-3}
\end{tabular}
\end{table}

\begin{table}[!htb] %
\caption{BBC News Codebook}
\vspace*{2.5mm}
\begin{tabular}{@{}lllll@{}}
\cmidrule(r){1-3}
\textbf{Code   Name} & \textbf{Original   Code Description} & \textbf{Revised   Code Description}     &  &  \\ \cmidrule(r){1-3}
business & business & Is this news story about business? &  &  \\ \\
entertainment        & entertainment                        & Is this news story about entertainment? &  &  \\ \\
politics & politics & Is this news story about politics? &  &  \\ \\
sport    & sport    & Is this news story about sports?   &  &  \\ \\
tech     & tech     & Is this news story about tech?     &  &  \\ \cmidrule(r){1-3}
\end{tabular}
\end{table}

\begin{longtable}{@{}lp{6.5cm}p{6.5cm}ll@{}}
\caption{Contrarian Claims Codebook}\\
\cmidrule(r){1-3}
\textbf{Code   Name} &
  \textbf{Original   Code Description} &
  \textbf{Revised   Code Description} &
   &
   \\* \cmidrule(r){1-3}
\endfirsthead
\endhead
1.1 &
  Ice/permafrost/snow cover isn't melting &
  Ice/permafrost/snow cover isn't melting &
   &
   \\ \\
1.2 &
  We're heading into an ice age/global cooling &
  We're heading into an ice age/global cooling &
   &
   \\ \\
1.3 &
  Weather is cold/snowing &
  Weather is cold/snowing &
   &
   \\ \\
1.4 &
  Climate hasn't warmed/changed over the last (few) decade(s) &
  Climate hasn't warmed/changed over the last (few) decade(s) &
   &
   \\ \\
1.5 &
  Oceans are cooling/not warming &
  Oceans are cooling/not warming &
   &
   \\ \\
1.6 &
  Sea level rise is exaggerated/not accelerating &
  Sea level rise is exaggerated/not accelerating &
   &
   \\ \\
1.7 &
  Extreme weather isn't increasing/has happened before/isn't   linked to climate change &
  Extreme weather isn't increasing/has happened before/isn't   linked to climate change &
   &
   \\ \\
1.8 &
  They changed the name from 'global warming' to 'climate change' &
  They changed the name from 'global warming' to 'climate change' &
   &
   \\ \\
2.1 &
  It's natural cycles/variation &
  It's natural cycles/variation &
   &
   \\ \\
2.2 &
  It's non-greenhouse gas human climate forcings (aerosols, land   use) &
  It's non-greenhouse gas human climate forcings (aerosols, land   use) &
   &
   \\ \\
2.3 &
  There's no evidence for greenhouse effect/carbon dioxide driving   climate change &
  There's no evidence for greenhouse effect/carbon dioxide driving   climate change &
   &
   \\ \\
2.4 &
  CO2 is not rising/ocean pH is not falling &
  CO2 is not rising/ocean pH is not falling &
   &
   \\ \\
2.5 &
  Human CO2 emissions are miniscule/not raising atmospheric CO2 &
  Human CO2 emissions are miniscule/not raising atmospheric CO2 &
   &
   \\ \\
3.1 &
  Climate sensitivity is low/negative feedbacks reduce warming &
  Climate sensitivity is low/negative feedbacks reduce warming &
   &
   \\ \\
3.2 &
  Species/plants/reefs aren't showing climate impacts yet/are   benefiting from climate change &
  Species/plants/reefs aren't showing climate impacts yet/are   benefiting from climate change &
   &
   \\ \\
3.3 &
  CO2 is beneficial/not a pollutant &
  CO2 is beneficial/not a pollutant &
   &
   \\ \\
3.4 &
  It's only a few degrees (or less) &
  It's only a few degrees (or less) &
   &
   \\ \\
3.5 &
  Climate change does not contribute to human conflict/threaten   national security &
  Climate change does not contribute to human conflict/threaten   national security &
   &
   \\ \\
3.6 &
  Climate change doesn't negatively impact health &
  Climate change doesn't negatively impact health &
   &
   \\ \\
4.1 &
  Climate policies (mitigation or adaptation) are harmful &
  Climate policies (mitigation or adaptation) are harmful &
   &
   \\ \\
4.2 &
  Climate policies are ineffective/flawed &
  Climate policies are ineffective/flawed &
   &
   \\ \\
4.3 &
  It's too hard to solve &
  It's too hard to solve &
   &
   \\ \\
4.4 &
  Clean energy technology/biofuels won't work &
  Clean energy technology/biofuels won't work &
   &
   \\ \\
4.5 &
  People need energy (e.g., from fossil fuels/nuclear) &
  People need energy (e.g., from fossil fuels/nuclear) &
   &
   \\ \\
5.1 &
  Climate-related science is uncertain / unsound / unreliable (data, methods \& models) &
  Climate-related science is uncertain / unsound / unreliable (data, methods \& models) &
   &
   \\ \\
5.2 &
  Climate movement is alarmist / wrong / political / biased / hypocritical (people or groups) &
  Climate movement is alarmist / wrong / political / biased / hypocritical (people or groups) &
   &
   \\ \\
5.3 &
  Climate change (science or policy) is a conspiracy (deception) &
  Climate change (science or policy) is a conspiracy (deception) &
   &
   \\ \\
0.0 &
  None of the above &
  None of the above &
   &
   \\* \cmidrule(r){1-3}
\end{longtable}

\pagebreak
\section{Supplemental Results}
\label{sec:detailed_results}

\begin{longtable}{@{}llcccccccccc@{}}
\caption{Detailed IRR Results}\\
\toprule
\multirow{3}{*}{\textbf{Dataset}} &
  \multirow{3}{*}{\textbf{Code}} &
  \multirow{3}{*}{\textbf{N}} &
  \multicolumn{3}{c}{\textbf{Code Count}} &
  \multicolumn{3}{c}{\textbf{Gwet's AC1}} & \\* \cmidrule(l){4-12} 
 &
   &
   &
  \multirow{2}{*}{\textbf{Original}} &
  \multirow{2}{*}{\textbf{Replicated}} &
  \multirow{2}{*}{\textbf{LLM}} &
  \textbf{Original-} &
  \textbf{Original-} &
  \textbf{Replicated-} & \\
 &
   &
   &
   &
   &
   &
  \textbf{Replicated} &
  \textbf{LLM} &
  \textbf{LLM} &
 \\* \midrule
\endfirsthead
\endhead
Trump   Tweets      & HSTG          & 100 & 12 & 13 & 57 & 0.96 & 0.18 & 0.19 \\[1mm]
Trump   Tweets      & ATSN          & 100 & 29 & 29 & 51 & 1    & 0.58 & 0.58 \\[1mm]
Trump   Tweets      & CRIT          & 100 & 39 & 26 & 30 & 0.73 & 0.76 & 0.73 \\[1mm]
Trump Tweets        & MEDI          & 100 & 7  & 7  & 6  & 1    & 0.96 & 0.96 \\[1mm]
Trump   Tweets      & FAMY          & 100 & 5  & 2  & 3  & 0.97 & 0.96 & 0.95 \\[1mm]
Trump   Tweets      & PLCE          & 100 & 4  & 4  & 4  & 1    & 0.98 & 0.98 \\[1mm]
Trump   Tweets      & MAGA          & 100 & 4  & 5  & 6  & 0.99 & 0.98 & 0.99 \\[1mm]
Trump   Tweets      & CAPT          & 100 & 26 & 28 & 48 & 0.93 & 0.36 & 0.36 \\[1mm]
Trump   Tweets      & INDV          & 100 & 42 & 41 & 57 & 0.79 & 0.5  & 0.48 \\[1mm]
Trump   Tweets      & MARG          & 100 & 4  & 1  & 2  & 0.97 & 0.94 & 0.97 \\[1mm]
Trump   Tweets      & INTN          & 100 & 24 & 19 & 15 & 0.86 & 0.81 & 0.86 \\[1mm]
Trump   Tweets      & PRTY          & 100 & 5  & 18 & 23 & 0.81 & 0.76 & 0.9  \\[1mm]
Trump   Tweets      & IMMG          & 100 & 7  & 6  & 8  & 0.99 & 0.97 & 0.95 \\[1mm]
Ukraine   Water     & env\_problems & 100 & 51 & 8  & 53 & 0.23 & 0.64 & 0.15 \\[1mm]
Ukraine   Water     & pollution     & 100 & 38 & 22 & 47 & 0.59 & 0.55 & 0.51 \\[1mm]
Ukraine   Water     & treatment     & 100 & 18 & 8  & 15 & 0.84 & 0.88 & 0.89 \\[1mm]
Ukraine   Water     & climate       & 100 & 19 & 17 & 9  & 0.97 & 0.87 & 0.84 \\[1mm]
Ukraine   Water     & biomonitoring & 100 & 8  & 36 & 8  & 0.51 & 0.86 & 0.48 \\[1mm]
BBC                 & sport         & 100 & 25 & 24 & 23 & 0.98 & 0.97 & 0.98 \\[1mm]
BBC                 & business      & 100 & 20 & 17 & 18 & 0.79 & 0.88 & 0.79 \\[1mm]
BBC                 & politics      & 100 & 27 & 30 & 36 & 0.71 & 0.84 & 0.75 \\[1mm]
BBC                 & entertainment & 100 & 17 & 17 & 14 & 0.97 & 0.96 & 0.96 \\[1mm]
BBC                 & tech          & 100 & 11 & 10 & 9  & 0.94 & 0.98 & 0.94 \\[1mm]
Contrarian   Claims & 1.1           & 100 & 3  & 3  & 7  & 1    & 0.96 & 0.96 \\[1mm]
Contrarian   Claims & 1.2           & 100 & 1  & 1  & 0  & 0.98 & 0.99 & 0.99 \\[1mm]
Contrarian   Claims & 1.3           & 100 & 0  & 1  & 1  & 0.99 & 0.99 & 0.98 \\[1mm]
Contrarian   Claims & 1.4           & 100 & 0  & 0  & 2  & 1    & 0.98 & 0.98 \\[1mm]
Contrarian   Claims & 1.5           & 100 & 0  & 0  & 0  & 1    & 1    & 1    \\[1mm]
Contrarian   Claims & 1.6           & 100 & 1  & 1  & 1  & 1    & 1    & 1    \\[1mm]
Contrarian   Claims & 1.7           & 100 & 2  & 0  & 2  & 0.98 & 0.96 & 0.98 \\[1mm]
Contrarian   Claims & 1.8           & 100 & 0  & 0  & 0  & 1    & 1    & 1    \\[1mm]
Contrarian   Claims & 2.1           & 100 & 6  & 7  & 8  & 0.97 & 0.93 & 0.92 \\[1mm]
Contrarian   Claims & 2.2           & 100 & 0  & 0  & 0  & 1    & 1    & 1    \\[1mm]
Contrarian   Claims & 2.3           & 100 & 3  & 2  & 9  & 0.99 & 0.93 & 0.92 \\[1mm]
Contrarian   Claims & 2.4           & 100 & 0  & 0  & 0  & 1    & 1    & 1    \\[1mm]
Contrarian   Claims & 2.5           & 100 & 0  & 0  & 2  & 1    & 0.98 & 0.98 \\[1mm]
Contrarian   Claims & 3.1           & 100 & 2  & 0  & 0  & 0.98 & 0.98 & 1    \\[1mm]
Contrarian   Claims & 3.2           & 100 & 0  & 0  & 0  & 1    & 1    & 1    \\[1mm]
Contrarian   Claims & 3.3           & 100 & 3  & 3  & 3  & 0.98 & 0.98 & 0.98 \\[1mm]
Contrarian   Claims & 3.4           & 100 & 0  & 0  & 0  & 1    & 1    & 1    \\[1mm]
Contrarian   Claims & 3.5           & 100 & 0  & 0  & 0  & 1    & 1    & 1    \\[1mm]
Contrarian   Claims & 3.6           & 100 & 0  & 0  & 0  & 1    & 1    & 1    \\[1mm]
Contrarian   Claims & 4.1           & 100 & 5  & 5  & 4  & 0.96 & 0.97 & 0.97 \\[1mm]
Contrarian   Claims & 4.2           & 100 & 1  & 2  & 3  & 0.97 & 0.96 & 0.97 \\[1mm]
Contrarian   Claims & 4.3           & 100 & 0  & 2  & 0  & 0.98 & 1    & 0.98 \\[1mm]
Contrarian   Claims & 4.4           & 100 & 0  & 1  & 1  & 0.99 & 0.99 & 1    \\[1mm]
Contrarian   Claims & 4.5           & 100 & 0  & 1  & 2  & 0.99 & 0.98 & 0.99 \\[1mm]
Contrarian   Claims & 5.1           & 100 & 10 & 12 & 10 & 0.95 & 0.93 & 0.93 \\[1mm]
Contrarian   Claims & 5.2           & 100 & 6  & 3  & 9  & 0.9  & 0.9  & 0.86 \\[1mm]
Contrarian   Claims & 5.3           & 100 & 0  & 2  & 0  & 0.98 & 1    & 0.98 \\[1mm]
Contrarian   Claims & 0.0           & 100 & 57 & 48 & 36 & 0.54 & 0.54 & 0.53 \\* \bottomrule
\end{longtable}

\end{document}